%% file: main.tex
\documentclass{article}

\usepackage[final]{nips_2017}

\usepackage[utf8]{inputenc} % allow utf-8 input
\usepackage[T1]{fontenc}    % use 8-bit T1 fonts
\usepackage{hyperref}       % hyperlinks
\usepackage{url}            % simple URL typesetting
\usepackage{booktabs}       % professional-quality tables
\usepackage{amsfonts}       % blackboard math symbols
\usepackage{nicefrac}       % compact symbols for 1/2, etc.
\usepackage{microtype}      % microtypography
\usepackage{amsmath}
\usepackage{amssymb}
\usepackage{graphicx}
\usepackage{soul}
\usepackage[usenames,dvipsnames,svgnames,table]{xcolor}

\usepackage[noend]{algpseudocode}
\usepackage{algorithm}
\newcommand{\pderiv}[2]{\frac{\partial #1}{\partial #2}}

\newcommand{\lossderiv}[1]{\frac{\partial \mathcal L}{\partial #1}}
\newcommand{\blue}[1]{\textcolor{blue}{#1}}

\usepackage{cancel}

\usepackage{hyperref} 

\usepackage{wrapfig}
\usepackage[framemethod=tikz]{mdframed}
\newenvironment{sidebar}[1][r]
  {\wrapfigure{#1}{0.5\textwidth}\mdframed[backgroundcolor=white!20,skipabove=0pt,skipbelow=0pt]}
  {\endmdframed\endwrapfigure}
\makeatletter
\newcommand{\mathleft}{\@fleqntrue\@mathmargin0pt}
\newcommand{\mathcenter}{\@fleqnfalse}
\makeatother

\title{Temporally Efficient Deep Learning with Spikes}

\author{
  Peter O'Connor\\
  QUVA Lab\\
  University of Amsterdam\\
  Amsterdam, Netherlands \\
  \texttt{peter.ed.oconnor@gmail.com} \\
  \And
  Efstratios Gavves \\
  QUVA Lab\\
  University of Amsterdam\\
  Amsterdam, Netherlands \\
  \texttt{e.gavves@uva.nl} \\
  \And
  Max Welling \\
  QUVA Lab\\
  University of Amsterdam\\
  Amsterdam, Netherlands \\
  \texttt{m.welling@uva.nl} \\
}

\begin{document}

\maketitle

\begin{abstract}
The vast majority of natural sensory data is temporally redundant. Video frames or audio samples which are sampled at nearby points in time tend to have similar values.  Typically, deep learning algorithms take no advantage of this redundancy to reduce computation.  This can be an obscene waste of energy.  We present a variant on backpropagation for neural networks in which computation scales with the rate of change of the data - not the rate at which we process the data.  We do this by having neurons communicate a combination of their state, and their temporal change in state.  Intriguingly, this simple communication rule give rise to units that resemble biologically-inspired leaky integrate-and-fire neurons, and to a weight-update rule that is equivalent to a form of Spike-Timing Dependent Plasticity (STDP), a synaptic learning rule observed in the brain.  We demonstrate that on MNIST and a temporal variant of MNIST, our algorithm performs about as well as a Multilayer Perceptron trained with backpropagation, despite only communicating discrete values between layers.

\end{abstract}

\section{Introduction}

Suppose we are trying to track objects in a scene.  A typical system used today would consist of sending camera-frames into a convolutional network which predicts bounding boxes.  Such a system may be trained by going over many hours of video with manually annotated bounding boxes, and learning to predict their locations.  This system has to execute a forward pass of a convolutional network at each iteration.  If we double the frame rate, we double the amount of computation, even if the contents of the video are mostly static.  Intuitively, it does not feel that this should be necessary.  Given the similarity between neighbouring frames of video, could we not reuse some of the computation from the last frame to update the bounding box inferences for the current frame?  Is it really necessary to recompute the entire network on each frame?

Many robotic systems consist of many sensors operating at wildly different frame rates.  Some ``neuromorphic'' sensors, such as the Dynamic Vision Sensor \cite{lichtsteiner2008128} have done away with the concept of frames altogether and instead send asynchronous ``events'' whenever the value of a pixel changes beyond some threshold.  It's not obvious, using current methods in deep learning, how we can efficiently integrate asynchronous sensory signals into a unified, trainable, latent representation, without recomputing the function of the network every time a new signal arrives.

There has been a lot of work on increasing the computational efficiency of neural networks by quantizing neural weights or activations (see Section \ref{sec:relwork}), but comparatively little work on exploiting redundancies in the data to reduce the amount of computation.  \cite{o2016sigma}, set out to exploit the temporal redundancy in video, by having neurons only send their quantized \textit{changes} in activation to downstream neurons, and having the downstream neurons integrate these changes.  This approach works for efficiently approximating the function of the network, but fails for training, because when the weights are changing with time, this approach (take the temporal difference, multiply by weights, temporally integrate) fails to reconstruct the correct activation for the next layer.  In other words, $\sum_{\tau=0}^t (x_{\tau}-x_{\tau-1})\cdot w_\tau \neq x_t\cdot w_t \;\text{unless}\; w_t=w_0 \forall t$.  Figure \ref{fig:why_kp} describes the problem visually.  In this paper, we correct for this by instead encoding a mixture of two components of the layers activation $x_t$: the \emph{proportional} component $k_p x_t$, and the \emph{derivative} component $k_d(x_t-x_{t-1})$. When we invert this encoding scheme, we get get a decoding scheme which corresponds to taking an exponentially decaying temporal average of past inputs.

Biological neurons tend to respond to a newly presented stimulus with a burst of spiking, which then decays to a slower baseline firing rate as the stimulus persists, and that neural membrane potentials can approximately be modeled as an exponentially decaying temporal average of past inputs.

\input{equation_sidebar}

\section{Methods}

We propose a coding scheme where neurons can represent their activations as a temporally sparse series of impulses.  The impulses from a given neuron encode a combination of the value and the rate of change of the neuron's activation.  

While our algorithm is designed to work efficiently with \emph{temporal data}, we do not aim to learn \emph{temporal sequences} in this work.  We aim to efficiently approximate a  function $y_t = f(x_t)$, where the current target $y_t$ is solely a function of the current input $x_t$, and not previous inputs $x_0...x_{t-1}$.   The temporal redundancy between neighbouring inputs $x_{t-1}, x_t$ will however be used to make our approximate computation of this function more efficient.

\subsection{Preliminary}

Throughout this paper we will use the notation $(f_3\circ f_2\circ f_1)(x) = f_3(f_2(f_1(x)))$ to denote function composition. We slightly abuse the notion of functions by allowing them to have an internal state which persists between calls. For example, we define the $\Delta$ function in Equation \ref{eq:delta} as being the difference between the inputs in two consecutive calls (where persistent variable $x_{last}$ is initialized to 0).  The $\Sigma$ function, defined in Equation \ref{eq:sigma}, returns a running sum of the inputs over calls.  So we can write, for example, that when our composition of functions $(\Sigma \circ \Delta)$ is called with a sequence of input variables $x_\tau: \tau = [1..t]$, then $(\Sigma \circ \Delta)(x_t) = x_t$, because $y_0 + (x_1-x_0) + (x_2-x_1) + ... + (x_t-x_{t-1}) |_{x_0=0, y_0=0} = x_t$. 

In general, when we write $y_t=f(x_t)$, where $f$ is a function with persistent state, it will be implied that we have previously called $f(x_\tau)$ for $\tau \in [1, .., t-1]$ in sequence.  Variable definitions that are used later will be highlighted in \blue{blue}. 

\subsection{PD Encoding}

Suppose a neuron has time-varying activation $x_\tau: \tau \in [1..t]$.  Taking inspiration from Proportional-Integral-Derivative (PID) controllers, we can ``encode'' this activation at each time step as a combination of its current activation and change in activation as $\blue{a_t \triangleq enc(x_t)=k_p x_t + k_d(x_t-x_{t-1})}$, (see Equation \ref{eq:enc}). The parameters $k_p$ and $k_d$ determine what portion of our encoding represents the value of the activation and the rate of change of that value, respectively.  In Section \ref{sec:tuning}, we will discuss the effect our choices for these parameters have on the network.  

To get our decoding formula, we can simply solve for $x_t$ as $x_t = \frac{a_t + k_d x_{t-1}}{k_p+k_d}$ (Equation \ref{eq:enc}), such that $(dec \circ enc)(x_t)=x_t$.  Notice that Equation \ref{eq:dec} corresponds to decaying the previous decoder state by some constant $k_d/(k_p+k_d)$ and then adding the input $a_t /(k_p+k_d)$.  We can expand this recursively to see that this corresponds to a temporal convolution $a * \kappa$ where $\kappa$ is a causal exponential kernel $\kappa_\tau = \left\{ \frac{1}{k_p+k_d}\left(\frac{k_d}{k_d+k_p}\right)^{\tau} \text{ if } \tau \geq0 \text{ otherwise } 0\right\}$. 

\subsection{Quantization}

Our motivation for the aforementioned encoding scheme is that we now want to quantize our signal into a sparse representation.  This will later be used to reduce computation.  We can quantize our signal  $a_t$ into a sparse, integer signal $\blue{s_t\triangleq Q(a_t)}$, where the quantizer Q is defined in Equation \ref{eq:quantize}.  Equation \ref{eq:quantize} implements a form of Sigma-Delta modulation, a method widely used in signal processing to approximately communicate signals at low bit-rates \citep{candy1962oversampling}.  We can show that that $Q(x_t) = (\Delta \circ R \circ \Sigma)(x_t)$ (See Supplementary Material Section \ref{app:sigmadelta}), where $\Delta \circ R \circ \Sigma$ indicates applying a temporal summation, a rounding, and a temporal difference, in series.  When $|a_t|\ll1 \forall t$, we can expect $s_t$ to consist of mostly zeros with a few 1's and -1's.  

We can now approximately reconstruct our original signal $x_t$ as $\blue{\hat x_t \triangleq dec (s_t)}$ by applying our decoder, as defined in Equation \ref{eq:dec}.  As our coefficients $k_p, k_d$ become larger, our reconstructed signal $\hat x_t$ should become closer to the original signal $x_t$.  We illustrate examples of encoded signals and their reconstructions for different $k_p$, $k_d$ in Figure \ref{fig:encdec}.  

\subsubsection{Special cases}

We can write compactly the entire reconstruction function as $\hat x = (dec \circ \Delta \circ R \circ \Sigma \circ enc)(x_t)$. 

\begin{figure}[t]
\includegraphics[width=.8\textwidth, trim={0 0.5cm 0 0},clip]{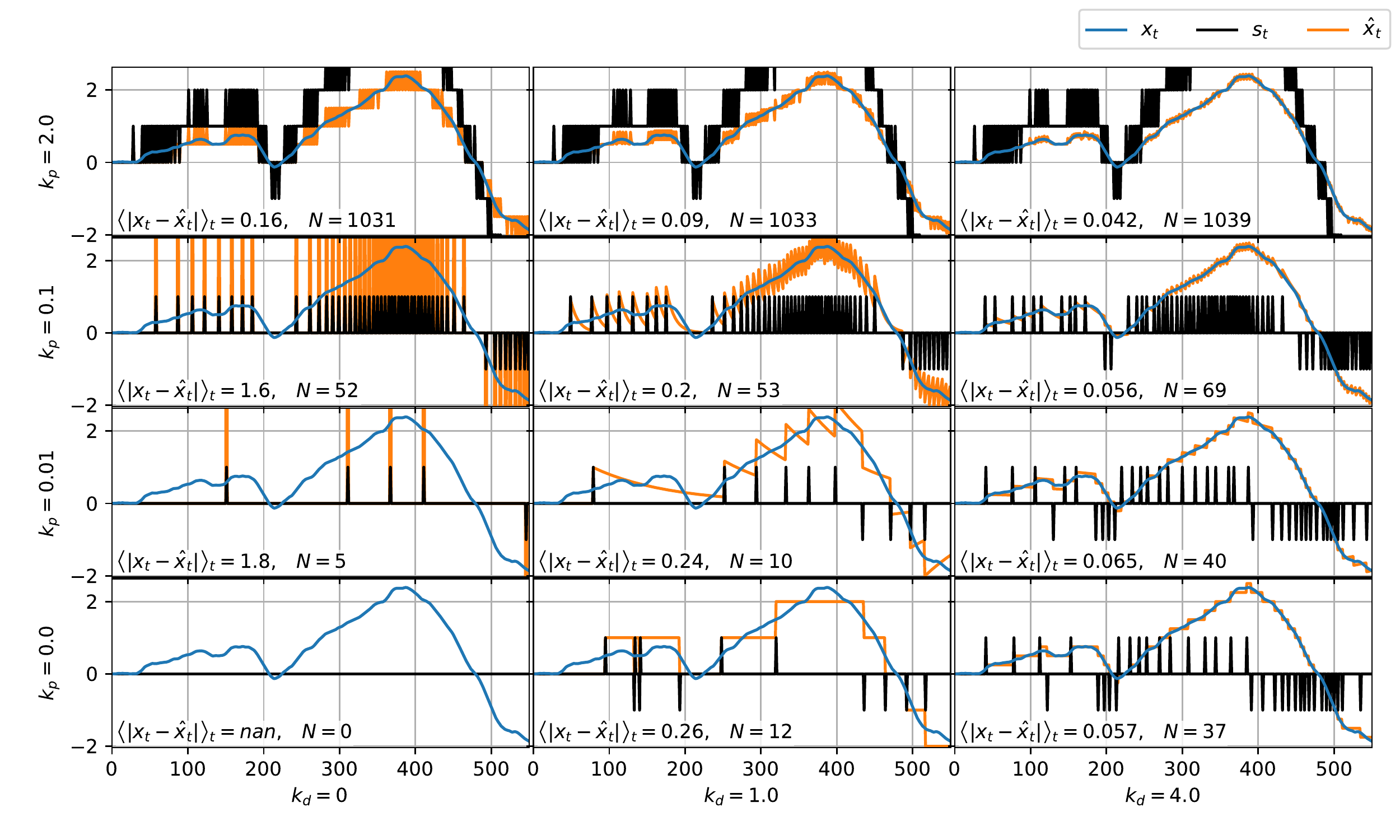}
\centering
\caption{An example signal $x_t$ (blue), encoded with $k_p$ varying across rows and $k_d$ varying across columns.  $s_t$ (black) is the quantized signal produced by the successive application of encoding (Equation \ref{eq:enc}) and quantization (Equation \ref{eq:quantize}.  $\hat x_t$ (orange) is the reconstruction of $x_t$ produced by applying Equation \ref{eq:dec} to $s_t$.  One might, after a careful look at this figure, ask why we bother with the proportional ($k_p$) term at all?  Figure 2 anticipates this question and answers it visually.  
}
\label{fig:encdec}
\end{figure}

\textbf{$k_p=0$}:  When $k_p=0$, we get $dec(x_t)=(k_d^{-1}\circ\Sigma)(x_t)$ and $enc(x_t)=(k_d\circ\Delta)(x_t)$, so our reconstruction reduces to $\hat x = (k_d^{-1}\circ\Sigma \circ \Delta \circ R \circ \Sigma \circ k_d\circ\Delta)(x_t)$. Because $\Sigma \circ k_d\circ\Delta$ all commute with one another, we can simplify this to $\hat x_t=(k_d^{-1}\circ R \circ k_d)(x_t)$.  so our decoded signal is $\hat x_t=round(x_t\cdot k_d)/k_d$, with no dependence on $x_{t-1}$.   This is visible in the bottom row of Figure \ref{fig:encdec}.  This was the encoding scheme used in \cite{o2016sigma}.

\textbf{$k_d=0$}: In this case, $dec(x_t)=k_p^{-1} x_t$ and $enc(x_t)=k_p x_t $ so our encoding-decoding process becomes $\hat x = (k_p^{-1}\circ\Delta \circ R \circ \Sigma \circ k_p)(x_t)$.  In this case neither our encoder nor our decoder have any memory, and we take not advantage of temporal redundancy.  

\begin{figure}[t]
\includegraphics[width=.8\textwidth]{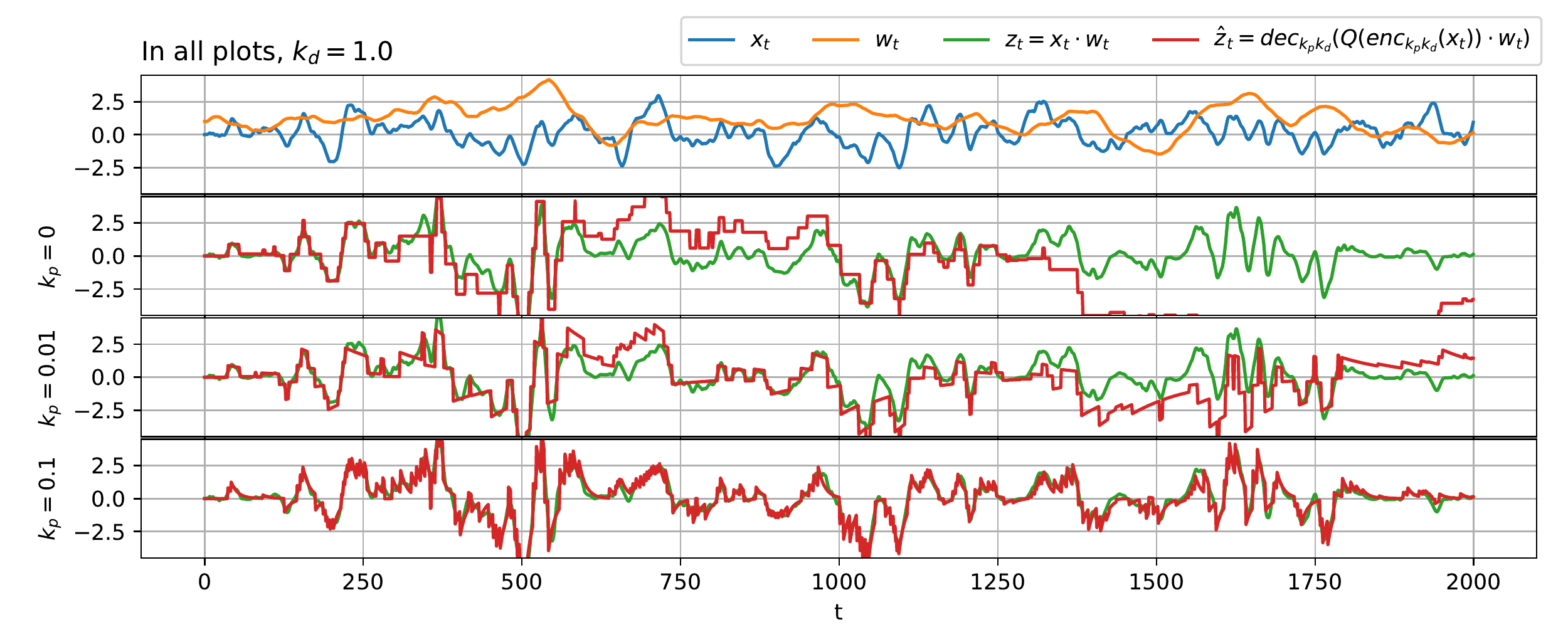}
\centering
\caption{The problem with \emph{only} sending changes in activation (i.e. $k_p=0$) is that during training, weights change over time.  In this example we generate random signals for a single scalar activation $x_t$ and scalar weight $w_t$.  We efficiently approximate $z_t$ with $\hat z_t$, as described in Section \ref{sec:innerprod}.  As the $w_t$ changes over time, our estimate $\hat z$ diverges from the correct value.  Introducing $k_p$ allows us to bring our reconstruction back in line with the correct signal.}
\label{fig:why_kp}
\end{figure}

\subsection{Sparse Communication Between Layers}
\label{sec:innerprod}

The purpose of our encoding scheme is to reduce computation by sparsifying communication between layers of a neural network.  Suppose we are trying to compute the pre-nonlinearity activation of the first hidden layer, $z_t\in \mathbb R^{d_{out}}$, given the input activation, $x_t \in \mathbb R^{d_{in}}$.  We approximate $z_t$ as:

\begin{equation}
\begin{split}
\blue{z_t \triangleq x_t \cdot w_t} \approx \hat x_t \cdot w_t \triangleq dec(Q(enc(x_t)))\cdot w_t \triangleq dec(s_t)\cdot w_t \approx \blue{dec(s_t\cdot w_t) \triangleq \hat z_t} \\
\text{where: } x_t, \hat x_t \in \mathbb R^{d_{in}}; s_t \in \mathbb I ^{d_{in}}; w\in \mathbb R^{d_{in}\times d_{out}}; z_t, \hat z_t \in \mathbb R^{d_{out}}
\label{eq:z-approx}
\end{split}
\end{equation}

The first approximation comes from the quantization (Q) of the encoded signal, and the second from the fact that the weights change over time, as explained in Figure \ref{fig:why_kp}.  The effects of these approximations are further explored in Section \ref{app:kscan} of the Supplementary Material.

Computing $z_t$ takes $d_{in}\cdot d_{out}$ multiplications and $(d_{in}-1)\cdot d_{out}$ additions.  The cost of computing $\hat z_t$, on the other hand, depends on the contents of $s_t$.  If the data is temporally redundant, $s_t\in \mathbb I^{d_{in}}$ should be sparse, with total magnitude $S\triangleq \sum_i|s_{t, i}|$.  $s_t$ can be decomposed into a sum of one-hot vectors $s_t=\sum_{n=1}^S sign(s_{t, i_n})\gamma_{i_n}:i_n\in[1..d_{in}]$ where $\gamma_{i_n}\in \mathbb I^{d_{in}}$ is a onehot vector with element $\gamma_{i_n}=1$.  The matrix product $s_t\cdot w$ can then be decomposed into a series of row additions:

\begin{equation}
\label{eq:dot}
s_t \cdot w = \left(\sum_{n=1}^{N}\text{sign}(s_{t,i_n}) \cdot \gamma_{i_n} \right)\cdot w 
= \sum_{n=1}^{N} \text{sign}(s_{t,i_n}) \gamma_{i_n}\cdot w
= \sum_{n=1}^{N} \text{sign}(s_{t,i_n}) \cdot w_{i_n, \cdot}
\end{equation}

If we include the encoding, quantization, and decoding operations, our matrix product takes a total of $2 d_{in} + 2 d_{out}$ multiplications, and $\sum_n|s_{t,n}|\cdot d_{out} + 3d_{in} + d_{out}$ additions.  Assuming the $\sum_n|s_{t,n}|\cdot d_{out}$ term dominates, we can say that the relative cost of computing $\hat z_t$ vs $z_t$  is:

\begin{equation}
\frac{cost(\hat z)}{cost(z)}\approx \frac{\sum_n|s_{t,n}|\cdot cost(add)}{d_{in}\cdot(cost(add) + cost(mult))}
\end{equation}

\subsection{A Neural Network}

We can implement this encoding scheme on every layer of a neural network.  Given a standard neural net $f_{nn}$ consisting of alternating linear ($\cdot w_l$) and nonlinear ($h_l$) operations, our network function $f_{pdnn}$ can then be written as:
% \begin{align}
% f_{nn}(x) &= (\cdot w_1; h_1; ... ; \cdot w_L;h_L)(x) \\
% f_{pdnn}(x) &= (enc_{k_p k_d};Q;\cdot w_1;dec_{k_p k_d};h_1; enc_{k_p k_d};Q; ... \cdot w_L ; h_L)(x)
% \end{align}
\begin{align}
f_{nn}(x) &= (h_L\circ \cdot w_L \circ ... \circ h_1 \circ w_1)(x) \\
f_{pdnn}(x) &= (h_L \circ w_L \circ Q_L \circ enc_L \circ ... \circ  h_1 \circ dec_1  \circ \cdot w_1 \circ Q_1 \circ enc_1)(x)
\end{align}

% $z_l\triangleq \{x\cdot w_1 \text{ if } l=1 \text{ otherwise } h_{l-1}(z_{l-1})\cdot w_l \}$   ... z_l for the original network

We can use the same approach to approximately calculate our gradients to use in training.  If we define our layer activations as $\hat z_l\triangleq (dec \circ \cdot w_l\circ Q \circ enc)(x) \text{ if } l=1 \text{ otherwise } (dec \circ \cdot w_l\circ Q \circ enc)(\hat z_{l-1})$, and $\mathcal L \triangleq \ell(f_{pdnn}(x), y)$, where $\ell$ is some loss function and $y$ is a target, we can backpropagate the approximate gradients as:

% \begin{align}
% \widehat{\pderiv{\mathcal L}{\hat z_l}} = 
% \begin{cases}
% \pderiv{\mathcal L}{z_L} &\text{if $l=L$} \\
%  \left(enc_{k_p k_d};Q;\cdot w_{l+1}^T;dec_{k_p k_d};\odot h_l'(\hat z_l)\right)(\widehat{\pderiv{\mathcal L}{\hat z_{l+1}}}) &\text{otherwise}
% \end{cases}
% \end{align}

\begin{align}
\widehat{\pderiv{\mathcal L}{\hat z_l}} = 
\begin{cases}
\pderiv{\mathcal L}{z_L} &\text{if $l=L$} \\
 \left( \odot h_l'(\hat z_l) \circ dec \circ \cdot w_{l+1}^T \circ Q \circ enc \right)(\widehat{\pderiv{\mathcal L}{\hat z_{l+1}}}) &\text{otherwise}
\end{cases}
\end{align}

On every layer of the forward and backward pass, our quantization scheme corrupts the signals that are being sent between layers.  Nevertheless we find experimentally that this does not matter much to the performance of the network.  

\subsection{Parameter Updates}
\label{sec:updates}

\begin{figure}[t]
\includegraphics[width=.7\textwidth]{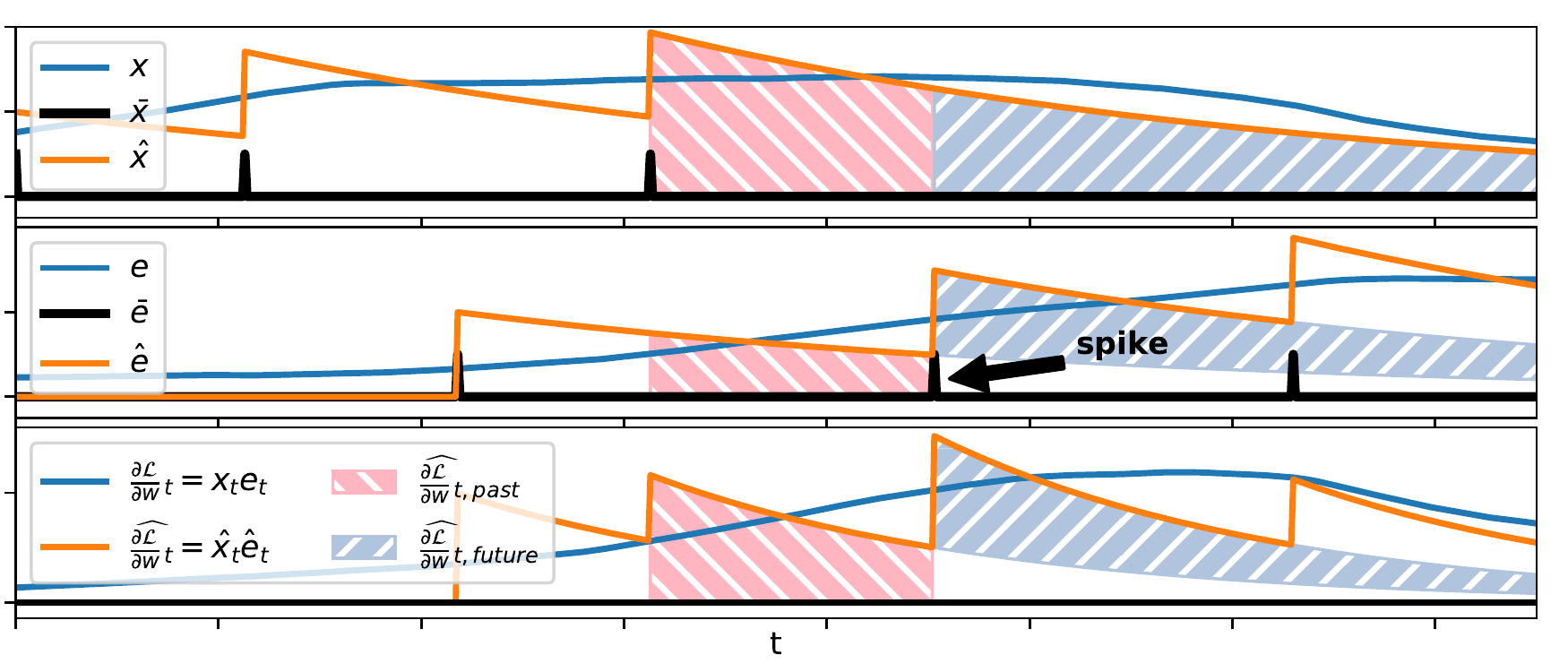}
\centering
\caption{A visualization of our efficient update schemes from Section \ref{sec:updates}.  \textbf{Top}: A scalar signal representing a presynaptic neuron activation $x_t=h_{l-1}(z_l-1)$, its quantized version, $\bar x_t = (Q\circ enc)(x_t)$, and its reconstruction $\hat x_t = dec(\bar x_t)$.  \textbf{Middle}: Another signal, representing the postsynaptic gradient of the error $e=\lossderiv{z_l}$, along with its quantized ($\bar e$) and reconstructed ($\hat e$) variants.  \textbf{Bottom}: The true weight gradient $\lossderiv{w_t}$, the reconstruction gradient $\hat{\lossderiv{w_t}}$.  At the time of the spike in $\bar e_t$, we have two schemes for efficiently computing the weight gradient that will be used to increment weight (see Section \ref{sec:updates}).  The \textit{past} scheme computes the area under $\hat x \cdot \hat e$ since the last spike, and the \textit{future} scheme computes the total future additional area due to the current spike.}
\label{fig:updates}
\end{figure}

There's no use having an efficient backward pass if the parameter updates aren't also efficient.  In a normal neural network trained with backpropagation and simple stochastic gradient descent, the parameter update for weight matrix $w$ has the form $w \leftarrow w -\eta \lossderiv{w}$ where $\eta$ is the learning rate.  If $w$ connects layer $l-1$ to layer $l$,  we can write $\lossderiv{w} = x_t \otimes e_t$ where $\blue{x_t\triangleq h_{l-1}(z_{l-1, t})\in \mathbb R^{d_{in}}}$ is the presynaptic activation, $\blue{e_t\triangleq \lossderiv{z_{l,t}}\in \mathbb R^{d_{out}}}$ is the postsynaptic (pre-nonlinearity) activation and $\otimes$ is the outer product.  So we pay $d_{in}\cdot d_{out}$ multiplications to update the parameters for each sample.

We want a more efficient way to compute this product, which takes advantage of the sparsity of our encoded signals to reduce computation.  We can start by applying our encoding-quantizing-decoding scheme to our input and error signals as $\blue{\bar x_t \triangleq (Q\circ enc)(x_t)\in \mathbb I^{d_{in}}}$ and $\blue{\bar e_t \triangleq (Q\circ enc)(e_t)\in \mathbb I^{d_{out}}}$, and approximate our true update update as $\blue{\widehat{\lossderiv{w}}_{recon,t}\triangleq\hat x_t \otimes \hat e_t}$ where $\blue{\hat x_t \triangleq dec(\bar x_t)}$ and $\blue{\hat e_t \triangleq dec(\bar e_t)}$.  This doesn't do any good by itself, because the update rule still is not sparse.  But, we can exactly compute the sum of this value over time using one of two sparse update schemes - \textit{past updates} and \textit{future updates} - which are depicted in Figure \ref{fig:updates}.  

\textbf{Past Updates}: For a given synapse $w_{i,j}$, if either the presynaptic neuron spikes ($\bar x_{t_i} \neq 0$) or the postsynaptic neuron spikes ($\bar e_{t_i} \neq 0$), we increment the $w_{i,j}$ by the total area under $\hat x_{\tau, i} \hat e_{\tau, j}$  since the last spike.  We can do this efficiently because between the current time and the time of the previous spike, $\hat x_{\tau, i} \hat e_{\tau, j}$ is a geometric sequence.  Given a known initial value $u$, final value $v$, and decay rate $r$, a geometric sequence sums to $\frac{u-v}{1-r}$.  The area calculated is shown in pink on the bottom row of Figure \ref{fig:updates}, and one algorithm to calculate it is in Equation \ref{eq:past}.

\textbf{Future Updates}: Another approach is to calculate the Present Value of the future area under the integral from the current spike.  This is depicted in the blue-gray area in Figure \ref{fig:updates}, and the formula is in Equation \ref{eq:future}. 

To simplify our expressions in the update algorithms, we re-parametrize our $k_p, k_d$ coefficients as $\blue{k_\alpha\triangleq= \frac{k_d}{k_p+k_d}}$, $\blue{k_\beta \triangleq \frac{1}{k_p+k_d}}$.

\input{algorithms}

\subsection{Relation to STDP}
\label{sec:stdp}
\begin{figure}[t]
\includegraphics[width=0.6\textwidth]{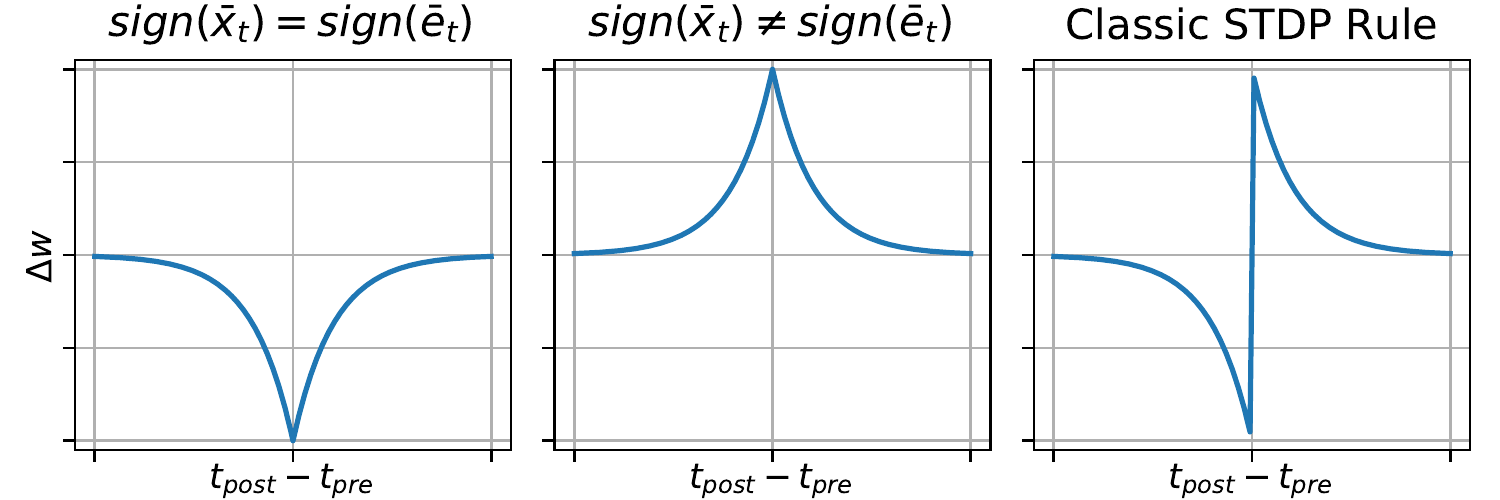}
\centering
\caption{\textbf{Left}: Our STDP rule, when both the input and error spikes have the same sign.  \textbf{Middle}: Our STDP rule, when the input and error spikes have opposite signs. \textbf{Right}: The classic STDP rule \cite{markram2012spike}, where the weight update is positive when a presynaptic spike preceeds a postsynaptic spike, and negative otherwise.}
\label{fig:stdp}
\end{figure}

An extremely attentive reader might have noted that Equation \ref{eq:future} has the form of an online implementation of Spike-Timing Dependent Plasticity (STDP).  STDP \citep{markram2012spike} emerged from neuroscience, where it was observed that synaptic weight changes appeared to be functions of the relative timing of pre- and post-synaptic spikes.  The empirically observed function usually has the double-exponential form seen on the rightmost plot of Figure \ref{fig:stdp}.  

Using the quantized input signal $\bar x$ and error signal $\bar e$, and their reconstructions $\hat x_t$ and $\hat e_t$ as defined in the last section, we define a causal convolutional kernel $\kappa_t = \left\{ k_\beta \left(k_\alpha\right)^{t} \text{ if } t\geq0 \text{ otherwise } 0\right\}$ and $g_t = \left\{  \kappa_t \text{ if } t\geq0 \text{ otherwise } \kappa_{-t}\right\} = k_\beta (k_\alpha)^{|t|} $ where $t\in I$.  The middle plot of Figure \ref{fig:stdp} is a plot of $g$.  We define our STDP update rule as:

\begin{align}
\widehat{\lossderiv{w}}_{t, STDP}= \left(\sum_{\tau=-\infty}^\infty \bar x_{t-\tau} g_\tau \right) \otimes \bar e_t
\end{align}
% &= \frac{1}{k_p^2+2k_p k_d}\sum_{t=1}^\infty ((\bar x * g) \odot \bar e)_t \label{eq:stdp}\\
% &= \frac{k_p+k_d}{k_p^2+2k_p k_d}\sum_{t=1}^\infty \left(\hat x_t \cdot \bar e_t + \bar x_t \cdot \hat e_t - \frac{\bar e_t \cdot \bar x_t}{k_p+k_d}\right)

We note that while our version of STDP has the same double-exponential form as the classic STDP rule observed in neuroscience \citep{markram2012spike}, we do not have the property that sign of the weight change depends on whether the presynaptic spike preceded the postsynaptic spike.  

In Section \ref{app:updates} in the supplementary material we show experimentally that while Equations $\widehat{\lossderiv{w}}_{recon}$, $\widehat{\lossderiv{w}}_{past}$, $\widehat{\lossderiv{w}}_{future}$, $\widehat{\lossderiv{w}}_{stdp}$ may all result in different updates at different times, the rules are equivalent in that for a given set of pre/post-synaptic spikes $\bar x, \bar e$, the cumulative sum of their updates over time converges exactly.

\subsection{Tuning $k_p$, $k_d$}
\label{sec:tuning}

The smaller the magnitude of a signal, the more severely distorted it is by our quantization-reconstruction scheme.  We can see that scaling a signal by K has the same effect on the quantized version of the signal, $s_t$, as scaling $k_p$ and $k_d$ by K: $s_t=(Q\circ enc_{k_p,k_d})(K x_t) = Q(k_p K x_t + k_d(K x_t - K x_{t-1})) = Q(Kk_p x_t + Kk_d(x_t-x_{t-1})) = (Q\circ enc_{Kk_p,Kk_d})(x_t)$. The fact that the reconstruction quality depends on the signal magnitude presents a problem when training our network, because the error gradients tend to change in magnitude throughout training (they start large, and become smaller as the network learns).  To keep our signal within the useful dynamic range of the quantizer, we apply simple scheme to heuristically adjust $k_p$ and $k_d$ for the forward and backward passes separately, for each layer of the network.  Instead of directly setting $k_p$, $k_d$ as hyperparameters, we fix the ratio $\blue{k_\alpha \triangleq \frac{k_d}{k_p+k_d}}$, and adapt the scale $\blue{k_\beta \triangleq \frac{1}{k_p+k_d}}$ to the magnitude of the signal.  Our update rule for $k_\beta$ is:

\begin{equation}
\begin{split}
\mu_t = (1-\eta_k) \mu_{t-1} + \eta_k \cdot |x_t|_{L_1} \\
k_\beta = k_\beta + \eta_k(k_\beta^{rel} \cdot \mu_t-k_\beta)
\end{split}
\end{equation}

Where $\eta_k$ is the scale-adaptation learning rate, $\mu_t$ is a rolling average of the $L_1$ magnitude of signal $x_t$, and $k_\beta^{rel}$ defines how coarse our quantization should be relative to the signal magnitude (higher means coarser).  We can recover $k_p, k_d$ for use in the encoders and decoders as $k_p = (1-k_\alpha)/k_\beta$ and $k_d = k_\alpha/k_\beta$.  In our experiments, we choose $\eta_k=0.001, k_\beta^{rel}=0.91, k_{alpha}=0.91$, and initialize $\mu_0=1$.

% \begin{figure}[t]
% \includegraphics[width=\textwidth]{pdnn_kscan2.pdf}
% \centering
% \caption{A demonstration of the effects of our choices of $k_p, k_d$ when weights are changing with time.  \textbf{Top Left:} A of snippet time-varying scalar signals $x_t, w_t \in \mathbb R$, and the $s_t=Q(enc_{k_p k_d} (x_t)) \in \{-1, 0, +1\}$.  \textbf{Bottom Left: } The true product of these signals (blue), compared to the reconstructed product when encoding and encoding+quantization are applied to x, respectively.  \textbf{Top Middle}: The Cosine similarity between the true signal $x \odot w$ and the non-quantized reconstructed signal $dec(enc(x)\odot w)$ (where $\odot$ denotes the elementwise product).  \textbf{Bottom Middle}: The Cosine similarity between the true signal and the quantized approximation $dec(Q(enc(x))\odot w)$.  \textbf{Bottom Left}: The number of spikes in the encoding $s_t$.
% }
% \label{fig:kscan}
% \end{figure}

\section{Experiments}
\label{sec:exp}
\begin{figure}
\begin{minipage}{.7\textwidth}
\includegraphics[width=1\textwidth]{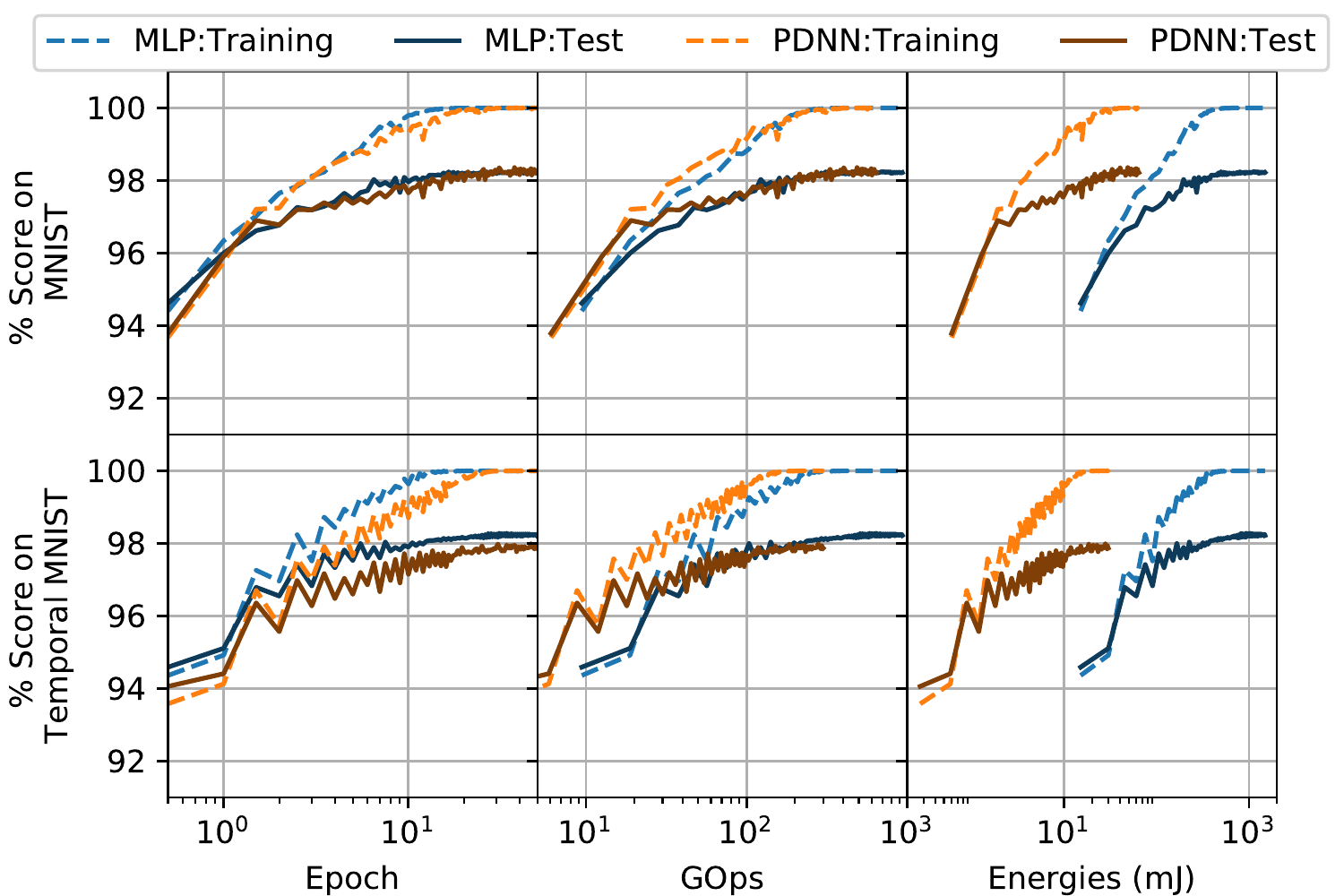}
	\caption{Top Row: Results on MNIST.  Bottom Row: Results on Temporal MNIST.  Left Column: the training and test scores as a function of epoch.  Middle: We now put the number of computational operations on the x-axis.  We see that as a result our PDNN shifts to the left.  Right: Because our network computes primarily with additions rather than multiplications.  When we multiply our operation counts with the estimates of \cite{horowitz20141} for the computational costs of arithmethic operations (0.1pJ for 32-bit fixed-point addition vs 32pJ for multiplication), we can see that our algorithm would be at an advantage on any hardware where arithmetic operations were the computational bottleneck.
    }
      \label{fig:mnist}
      
\end{minipage}
\hspace{.5cm}
\begin{minipage}{.25\textwidth}
      \centering
      \includegraphics[width=\textwidth, trim={0 4.9cm 0 0},clip]{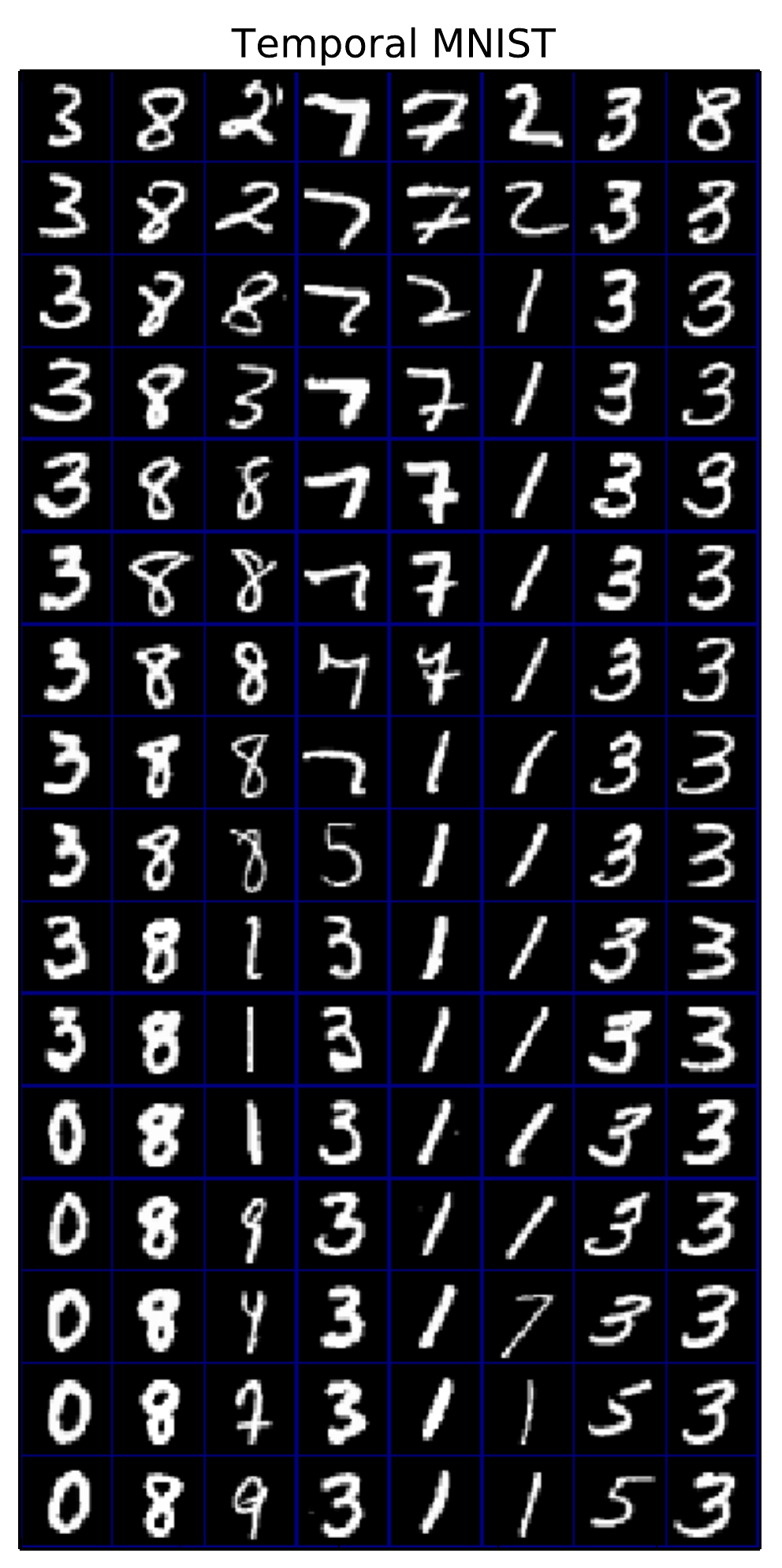}
      \caption{Some samples from the Temporal-MNIST dataset.  Each column shows a snippet of adjacent frames.}
      \label{fig:temporal-mnist}
\end{minipage}
\end{figure}

To evaluate our network's ability to learn, we run it on the standard MNIST dataset, as well as a variant we created called ``Temporal MNIST''.  Temporal MNIST is simply a reshuffling of the MNIST dataset so that so that similar inputs (in terms of L2-pixel distance), are put together.  Figure \ref{fig:temporal-mnist} shows several snippets of consecutive frames in the temporal MNIST dataset.  We compare our Proportional-Derivative Net against a conventional Multi-Layer Perceptron with the same architecture (one hidden layer of 200 ReLU hidden units and a softmax output).  The results are shown in Figure \ref{fig:mnist}.  Somewhat surprisingly, our predictor slightly outperformed the MLP, getting 98.36\% on the test set vs 98.25\% for the MLP.  We assume this improvement is due to the regularizing effect of the quantization.  On Temporal MNIST, our network was able to converge with less computation than it required for MNIST (It used $32\cdot10^{12}$ operations for MNIST vs $15\cdot 10^{12}$ for Temporal MNIST), but ended up with a slightly worse test score when compared with the MLP (the PDNN got 97.99\% vs 98.28\% for the MLP).  It's not clear why our network appeared to achieve a slightly worse score on temporal data.  This will be a subject for future investigation.

\section{Related Work}
\label{sec:relwork}
There has been sparse but interesting work on merging the notions of spiking neural networks and deep learning.  \cite{diehl2015fast} found a way to efficiently map a trained neural network onto a spiking network.  \cite{lee2016training} devised a method for training spiking of integrate-and-fire spiking neurons with backpropagation - though their neurons did not send a temporal difference of their activations. \cite{o2016deep} created a method for training event-based neural networks - but their method took no advantage of temporal redundancy in the data.  \cite{DBLP:journals/corr/BinasIP16} and \citep{o2016sigma} both took the approach of sending quantized temporal changes reduce computation on temporally redundant data, but their schemes could not be used to train a neural network.  \cite{bohte2000spikeprop} showed how could apply backpropagation for training spiking neural networks, but it was not obvious how to apply the method to non-spiking data.  \cite{zambrano2016fast} developed a spiking network with an adaptive scale of quantization (which bears some resemblance to our tuning scheme described in Section \ref{sec:tuning}), and show that the spiking mechanism is a form of Sigma-Delta modulation, which we also use here.  \cite{DBLP:journals/corr/CourbariauxBD15} showed that neural networks could be trained with binary weights and activations (we just quantize activations).  \cite{bengio2015objective} found a connection between the classic STDP rule (Figure \ref{fig:stdp}, right) and optimizing a dynamical neural network, although the way they arrived at an STDP-like rule was quite different from ours.  

\section{Discussion}

We set out with the objective of reducing the computation in deep networks by taking advantage of temporal redundancy in data.  We described a simple rule (Equation \ref{eq:enc}) for sparsifying the communication between layers of a neural network by having our neurons communicate a combination of their temporal change in activation, and the current value of their activation.  We show that it follows from this scheme that neurons should behave as leaky integrators (Equation \ref{eq:dec}).  When we quantize our neural activations with Sigma-Delta modulation, a common quantization scheme in signal processing, we get something resembling a leaky integrate-and-fire neuron.  We derive efficient update rules for the weights of our network, and show these to be equivalent to a form of STDP - a learning rule first observed in neuroscience.  Finally, we train our network, verify that it does indeed compute more efficiently on temporal data, and show that it performs about as well as a traditional deep network of the same architecture, but with significantly reduced computation.

Code is available at \href{https://github.com/petered/pdnn}{github.com/petered/pdnn}.

\subsubsection*{Acknowledgments}
This work was supported by Qualcomm, who we'd like to thank for sharing their past work with us.  In addition, we'd like to thank our colleagues, especially Matthias Reisser and Changyong Oh, for some very useful discussions which contributed to this work.

\bibliographystyle{plainnat} 
\bibliography{mybib}

\clearpage
\appendix
\section{Sigma-Delta Unwrapping}
\label{app:sigmadelta}

Here we show that $Q = \Delta \circ R \circ \Sigma$, where $Q, \Delta, R, \Sigma$ are defined in Equations \ref{eq:quantize}, \ref{eq:sigma}, \ref{eq:round}, \ref{eq:delta}, respectively.

From Equation \ref{eq:quantize} (Q) we can see that
\begin{equation*}\begin{aligned}
& y_t \leftarrow round(x_t + \phi_{t-1}) \in \mathbb I\\
&\phi_t \leftarrow \phi_{t-1} + x_t - y_t \in \mathbb R
\end{aligned}\end{equation*}

Now we can unroll for $y_t$ and observe use the fact that if $s\in \mathbb I$ then $round(a+s)=round(a)+s$, to say:

\begin{equation}\begin{aligned}
y_t &= round(x_t + \phi_{t-1}) \\
&= round(x_t + \phi_{t-2} + x_{t-1} - y_{t-1}) \\
&= round(x_t + x_{t-1} + \phi_{t-2}) - y_{t-1} \\
&= round(x_t + x_{t-1} + \phi_{t-2}) - round(x_{t-1} + \phi_{t-2}) \\
&= \left(round(\sum_{\tau=1}^t x_\tau + \cancelto{0}{\phi_0}) - \sum_{\tau=0}^{t-2} y_{\tau}\right) - \left(round(\sum_{\tau=1}^{t-1} x_\tau + \cancelto{0}{\phi_0}) - \sum_{\tau=0}^{t-2} y_{\tau}\right) \\
&= round(\sum_{\tau=1}^t x_\tau) - round(\sum_{\tau=1}^{t-1} x_\tau)
\end{aligned}\end{equation}

At which point it is clear that Q is identical to a successive application of a temporal summation, a rounding, and a temporal difference.  That is why we say $Q = \Delta \circ R \circ \Sigma$.

\section{Scanning the K-space}
\label{app:kscan}
Equation \ref{eq:z-approx} shows how we make two approximations when approximating $z_t = x_t \cdot w_t$ with $\hat z_t = (dec \circ w \circ Q\circ enc)(x_t)$.  The first is the ``nonstationary weight'' approximation - arising from the fact that w changes in time, the second is the ``quantization'' approximation, arising from the quantization of x.  Here do a small experiment in which we multiply a time-varying scalar signal $x_t$ with a time-varying weight $w_t$ for many different values of $k_p, k_d$ to understand the effects of $k_p, k_d$ on our approximation error.

\begin{figure}[H]
\includegraphics[width=.9\textwidth]{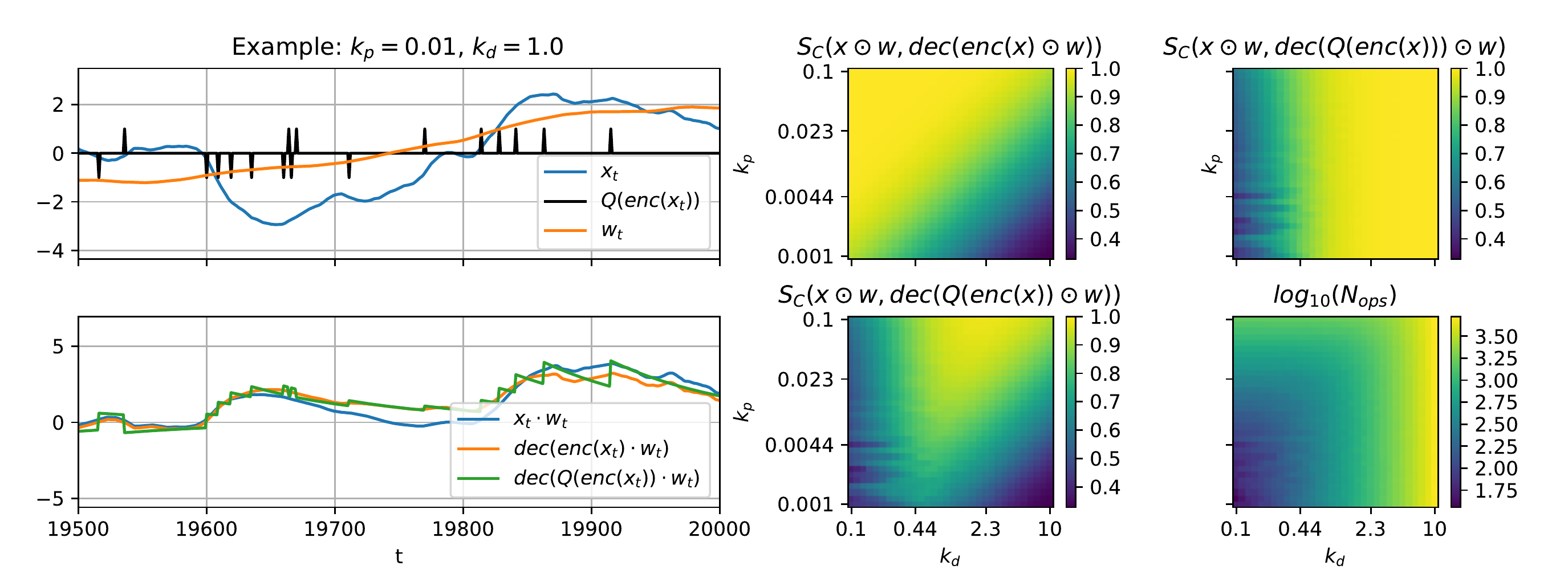}
\centering
\caption{\textbf{Top Left}: A time varying signal $x_t$, the quantized signal $Q(enc(x_t))$, and the time-varying ``weight'' $w_t$.  \textbf{Bottom Left}: Compare the true product of these signals $x_t\cdot w_t$ with the $dec(enc(x_t)\cdot w_t)$, which shows the effects of the non-stationary weight approximation, and $dec(Q(enc(x_t))\cdot w)$ which shows both approximations.  \textbf{Top Middle}: The Cosine distance between the ``true'' signal $x\odot w$ and the approximation due to the nonstationary w, scanned over a grid of $k_p, k_d$ values.  \textbf{Top Right}: The cosine distance between the ``true'' signal and the approximation due to the quantization of x.  \textbf{Bottom Middle}: The Cosine Distance between the ``true'' signal and the full approximation described in Equation \ref{eq:z-approx}.  This shows why we need both $k_p$ and $k_d$ to be nonzero.  \textbf{Bottom Right}: The Number of weight-lookups required for the to compute the full approximation. $dec(Q(enc(x))\odot w)$.}
\label{fig:kscan}
\end{figure}

\section{All roads lead to Rome}
\label{app:updates}

In Section \ref{sec:updates} and \ref{sec:stdp}, we described 4 different update rules, and stated that while they do not necessarily produce the same updates at the same times, they produce the same result in the end.  Here we demonstrate this empirically.  We generate two random spike-trains representing the input and the error signal to a single synapse.  The plot on the bottom shows our weight as a function of time as it drifts from its initial value.  

\begin{figure}[H]
\includegraphics[width=\textwidth]{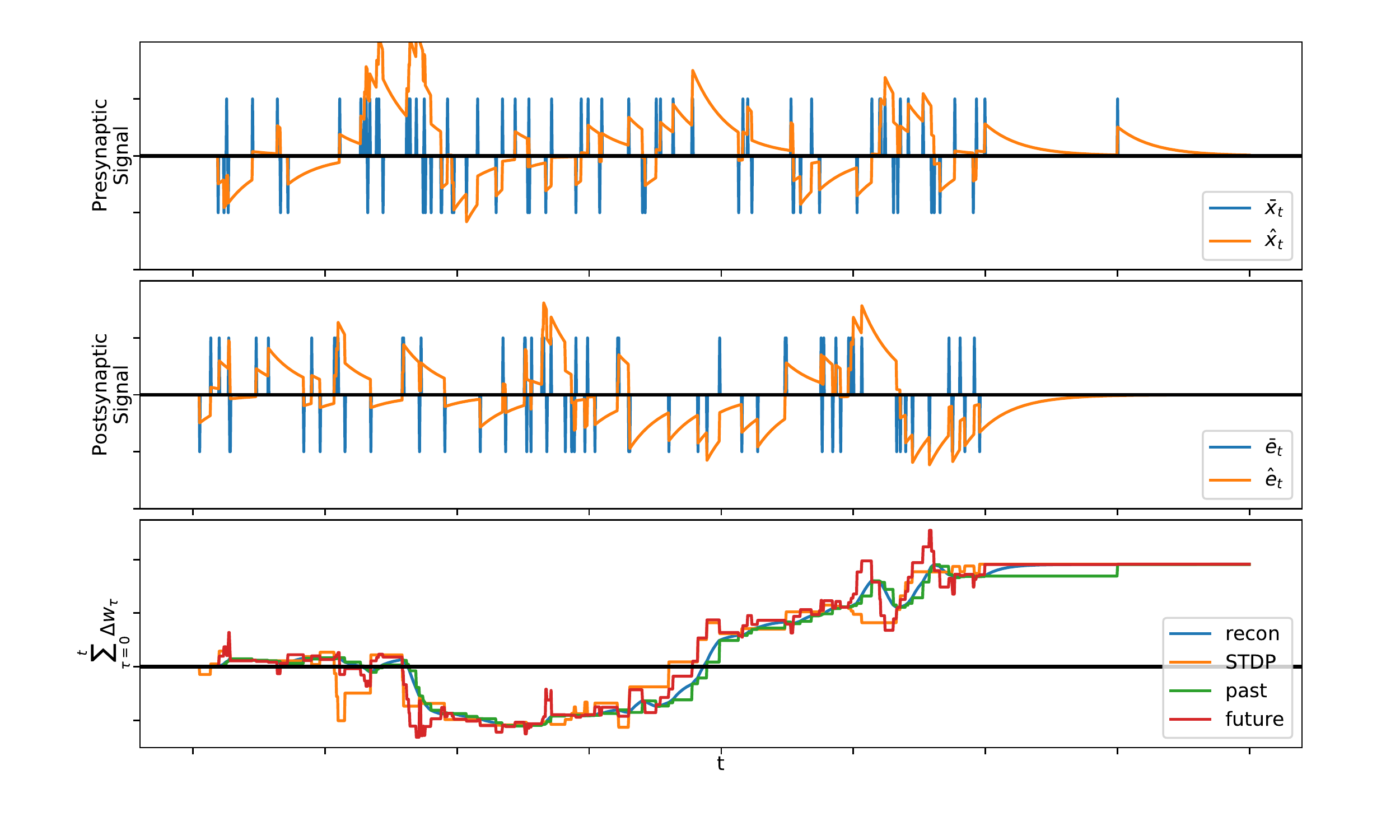}
\centering
\caption{A comparison of our different update methods.  \textbf{Top}: A randomly generated presynaptic quantized signal $\bar x$, along with its reconstruction $\hat x$.  \textbf{Middle}: A randomly generated postsynaptic quantized error signal $\bar e$, along with its reconstruction $\hat e$.  \textbf{Bottom}: The cumulative weight update arising from our four updates methods.  "\textit{recon}" is just $\sum_{\tau=1}^t \hat x_\tau \hat e_\tau$, ``\textit{past}'' and ``\textit{future}'' are described in Section \ref{sec:updates} and ``\textit{STDP}'' is described in Section \ref{sec:stdp}
}
\label{fig:updates_eq}
\end{figure}

\end{document}

%% file: equation_sidebar.tex
\begin{sidebar}
\mathleft
\parskip=-5pt

\begin{equation}\begin{aligned}
\Delta: & x\mapsto y; \;\;\text{Persistent:}\;\; x_{last} \leftarrow 0 \\
& y \leftarrow x-x_{last} \\
& x_{last} \leftarrow x
\end{aligned}\label{eq:delta}\end{equation}

\hrulefill

\begin{equation}\begin{aligned}
\Sigma: &x\mapsto y; \;\;\text{Persistent:}\;\; y \leftarrow 0 \\
& y \leftarrow y+x\\
\end{aligned}\label{eq:sigma}\end{equation}

\hrulefill

\begin{equation}\begin{aligned}
Q: & x \mapsto y; \;\;\text{Persistent:}\;\;\phi \leftarrow 0 \\
&\phi' \leftarrow \phi + x \\
&y \leftarrow round(\phi') \\
&\phi \leftarrow \phi' - y
\end{aligned}\label{eq:quantize}\end{equation}

\hrule

\begin{equation}\begin{aligned}
enc: & x \mapsto y; \;\;\text{Persistent:}\;\;x_{last} \leftarrow 0 \\
&y \leftarrow k_p  x + k_d (x-x_{last}) \\
&x_{last} \leftarrow x
\end{aligned}\label{eq:enc}\end{equation}

\hrule

\begin{equation}\begin{aligned}
dec: & x \mapsto y;\;\text{Persistent:}\;\;y \leftarrow 0 \\
&y \leftarrow \frac{x + k_d y}{k_p+k_d} 
\end{aligned}\label{eq:dec}\end{equation}

\hrule

\begin{equation}\begin{aligned}
R: & x \mapsto round(x)
\end{aligned}\label{eq:round}\end{equation}

\mathcenter
\end{sidebar}

%% file: algorithms.tex
    \begin{tabular}{|l|l|}   
    
      \hline
      \begin{minipage}[t]{.45\textwidth}
      \begin{equation}\begin{aligned}
      p&ast :(\bar x\in \mathbb I^{d_{in}}, \bar e\in \mathbb I^{d_{out}}) \mapsto \widehat{\lossderiv w}_{past} \\
      &\text{Persistent: }w, u \in \mathbb R^{d_{in}\times d_{out}},\\ 
      & \qquad\qquad \hat x\leftarrow 0^{d_{in}}, \hat e \leftarrow 0^{d_{out}} \\
      &i \leftarrow \bar x \neq 0 ,\quad j \leftarrow \bar e \neq 0 \\
      &\hat x \leftarrow k_\alpha \hat x \quad,\quad  \hat e \leftarrow k_\alpha \hat e \\
      &v \leftarrow \hat x_{i} \otimes \hat e_j \in \mathbb R ^{\sum_{i'}[\bar x_{i'}\neq 0] \times  \sum_{j'}[\bar e_{j'}\neq 0]}\\
      &\widehat{\lossderiv w}_{past, i, j} \leftarrow \frac{u_{i,j}-v}{1-k_\alpha^2} \\
      &\hat x \leftarrow \hat x + k_\beta \bar x,\quad\hat e \leftarrow \hat e + k_\beta \bar e \\
      &u_{i,j} \leftarrow v \\
      \end{aligned}\label{eq:past}\end{equation}
      \end{minipage}
      &
      \begin{minipage}[t]{.45\textwidth}
      \begin{equation}\begin{aligned}
      f&uture :(\bar x\in \mathbb I^{d_{in}}, \bar e\in \mathbb I^{d_{out}}) \mapsto \widehat{\lossderiv w}_{future} \\
      &\text{Persistent: }w \in \mathbb R^{d_{in}\times d_{out}},\\ 
      & \qquad\qquad \hat x \leftarrow 0^{d_{in}}, \hat e \leftarrow 0^{d_{out}} \\
      &\hat x \leftarrow k_\alpha \hat x  \\
      & \hat e \leftarrow k_\alpha \hat e + k_\beta \bar e \\
      &\widehat{\lossderiv w}_{future}\leftarrow \frac{\bar x \otimes \hat e + \hat x \otimes \bar e}{k_\alpha^2-1}\\
      &\hat x \leftarrow \hat x+k_\beta \bar x \\
      \end{aligned}\label{eq:future}\end{equation}
      \end{minipage}
   \\
   \\
   \hline
   \end{tabular}

%% file: main.bbl
\begin{thebibliography}{13}
\providecommand{\natexlab}[1]{#1}
\providecommand{\url}[1]{\texttt{#1}}
\expandafter\ifx\csname urlstyle\endcsname\relax
  \providecommand{\doi}[1]{doi: #1}\else
  \providecommand{\doi}{doi: \begingroup \urlstyle{rm}\Url}\fi

\bibitem[Bengio et~al.(2015)Bengio, Mesnard, Fischer, Zhang, and
  Wu]{bengio2015objective}
Yoshua Bengio, Thomas Mesnard, Asja Fischer, Saizheng Zhang, and Yuhai Wu.
\newblock An objective function for stdp.
\newblock \emph{arXiv preprint arXiv:1509.05936}, 2015.

\bibitem[Binas et~al.(2016)Binas, Indiveri, and
  Pfeiffer]{DBLP:journals/corr/BinasIP16}
Jonathan Binas, Giacomo Indiveri, and Michael Pfeiffer.
\newblock Deep counter networks for asynchronous event-based processing.
\newblock \emph{CoRR}, abs/1611.00710, 2016.
\newblock URL \url{http://arxiv.org/abs/1611.00710}.

\bibitem[Bohte et~al.(2000)Bohte, Kok, and La~Poutr{\'e}]{bohte2000spikeprop}
Sander~M Bohte, Joost~N Kok, and Johannes~A La~Poutr{\'e}.
\newblock Spikeprop: backpropagation for networks of spiking neurons.
\newblock In \emph{ESANN}, pages 419--424, 2000.

\bibitem[Candy and Temes(1962)]{candy1962oversampling}
James~C Candy and Gabor~C Temes.
\newblock \emph{Oversampling delta-sigma data converters: theory, design, and
  simulation}.
\newblock University of Texas Press, 1962.

\bibitem[Courbariaux et~al.(2015)Courbariaux, Bengio, and
  David]{DBLP:journals/corr/CourbariauxBD15}
Matthieu Courbariaux, Yoshua Bengio, and Jean{-}Pierre David.
\newblock Binaryconnect: Training deep neural networks with binary weights
  during propagations.
\newblock \emph{CoRR}, abs/1511.00363, 2015.
\newblock URL \url{http://arxiv.org/abs/1511.00363}.

\bibitem[Diehl et~al.(2015)Diehl, Neil, Binas, Cook, Liu, and
  Pfeiffer]{diehl2015fast}
Peter~U Diehl, Daniel Neil, Jonathan Binas, Matthew Cook, Shih-Chii Liu, and
  Michael Pfeiffer.
\newblock Fast-classifying, high-accuracy spiking deep networks through weight
  and threshold balancing.
\newblock In \emph{2015 International Joint Conference on Neural Networks
  (IJCNN)}, pages 1--8. IEEE, 2015.

\bibitem[Horowitz(2014)]{horowitz20141}
Mark Horowitz.
\newblock 1.1 computing's energy problem (and what we can do about it).
\newblock In \emph{2014 IEEE International Solid-State Circuits Conference
  Digest of Technical Papers (ISSCC)}, pages 10--14. IEEE, 2014.

\bibitem[Lee et~al.(2016)Lee, Delbruck, and Pfeiffer]{lee2016training}
Jun~Haeng Lee, Tobi Delbruck, and Michael Pfeiffer.
\newblock Training deep spiking neural networks using backpropagation.
\newblock \emph{arXiv preprint arXiv:1608.08782}, 2016.

\bibitem[Lichtsteiner et~al.(2008)Lichtsteiner, Posch, and
  Delbruck]{lichtsteiner2008128}
Patrick Lichtsteiner, Christoph Posch, and Tobi Delbruck.
\newblock A 128$\times$ 128 120 db 15 $\mu$s latency asynchronous temporal
  contrast vision sensor.
\newblock \emph{Solid-State Circuits, IEEE Journal of}, 43\penalty0
  (2):\penalty0 566--576, 2008.

\bibitem[Markram et~al.(2012)Markram, Gerstner, and
  Sj{\"o}str{\"o}m]{markram2012spike}
Henry Markram, Wulfram Gerstner, and Per~Jesper Sj{\"o}str{\"o}m.
\newblock Spike-timing-dependent plasticity: a comprehensive overview.
\newblock \emph{Frontiers in synaptic neuroscience}, 4, 2012.

\bibitem[O'Connor and Welling(2016{\natexlab{a}})]{o2016deep}
Peter O'Connor and Max Welling.
\newblock Deep spiking networks.
\newblock \emph{arXiv preprint arXiv:1602.08323}, 2016{\natexlab{a}}.

\bibitem[O'Connor and Welling(2016{\natexlab{b}})]{o2016sigma}
Peter O'Connor and Max Welling.
\newblock Sigma delta quantized networks.
\newblock \emph{arXiv preprint arXiv:1611.02024}, 2016{\natexlab{b}}.

\bibitem[Zambrano and Bohte(2016)]{zambrano2016fast}
Davide Zambrano and Sander~M Bohte.
\newblock Fast and efficient asynchronous neural computation with adapting
  spiking neural networks.
\newblock \emph{arXiv preprint arXiv:1609.02053}, 2016.

\end{thebibliography}
